\documentclass[11pt]{article}

\usepackage[final]{acl}

\usepackage{times}
\usepackage{latexsym}

\usepackage{amsmath,amsfonts,bm}

\def\eqref#1{equation~\ref{#1}}

\def\1{\bm{1}}

\DeclareMathAlphabet{\mathsfit}{\encodingdefault}{\sfdefault}{m}{sl}
\SetMathAlphabet{\mathsfit}{bold}{\encodingdefault}{\sfdefault}{bx}{n}

\usepackage[T1]{fontenc}

\usepackage[utf8]{inputenc}
\usepackage{hyperref}
\usepackage{float}
\usepackage{microtype}
\usepackage{balance}

\usepackage{inconsolata}

\usepackage{graphicx}
 \usepackage{amsmath} 
\definecolor{kleinblue}{RGB}{0, 47, 167}

\usepackage[most]{tcolorbox}
\usepackage{xcolor}
\usepackage{tabularx} 
\usepackage{hyperref}
\usepackage{framed} 
\usepackage{cuted} 
\usepackage{enumitem}

\definecolor{lightpastelblue}{RGB}{210, 230, 255}
\definecolor{lightpastelgreen}{RGB}{220, 240, 220}
\definecolor{lightpastelyellow}{RGB}{255, 255, 220}
\definecolor{lightpastelpink}{RGB}{255, 230, 235}

\newcounter{conclusion}
\renewcommand{\theconclusion}{\arabic{conclusion}}

\newtcolorbox{conclusionbox}[1][]{%
    enhanced,
    colframe=lightpastelblue!70!white,
    colback=lightpastelblue!10,
    coltitle=black,
    fonttitle=\bfseries,
    title=Conclusion~\theconclusion,
    arc=10pt,
    boxrule=1.2pt,
    left=10pt,
    right=10pt,
    top=10pt,
    bottom=10pt,
    titlerule=1pt,
    title style={fill=lightpastelblue!25},
    before upper={\stepcounter{conclusion}}, 
    #1 
}
\hypersetup{
    colorlinks=true,
    linkcolor=kleinblue,
    filecolor=kleinblue,
    citecolor=kleinblue,
    urlcolor=kleinblue,
    }
\usepackage{url}
\usepackage{graphicx}
\usepackage{booktabs}
\usepackage{multirow}
\usepackage{pifont}
\usepackage{xcolor}
\usepackage{framed} 
\usepackage{amsmath}
\usepackage{subcaption}   
\usepackage{caption}

\usepackage{tikz}
\usetikzlibrary{arrows.meta,calc,positioning,fit,decorations.markings,backgrounds}

\definecolor{darkgreen}{RGB}{0,128,0}
\definecolor{darkred}{RGB}{139,0,0}
\definecolor{gray}{rgb}{0.5, 0.5, 0.5}
\definecolor{darkyellow}{rgb}{1,0.75,0.1}

\newcommand{\cmark}{\ding{51}} 
\newcommand{\xmark}{\ding{55}} 
\newcommand{\abmark}{\ding{162}} 

\usepackage{xcolor}

\title{\includegraphics[width=0.04\linewidth]{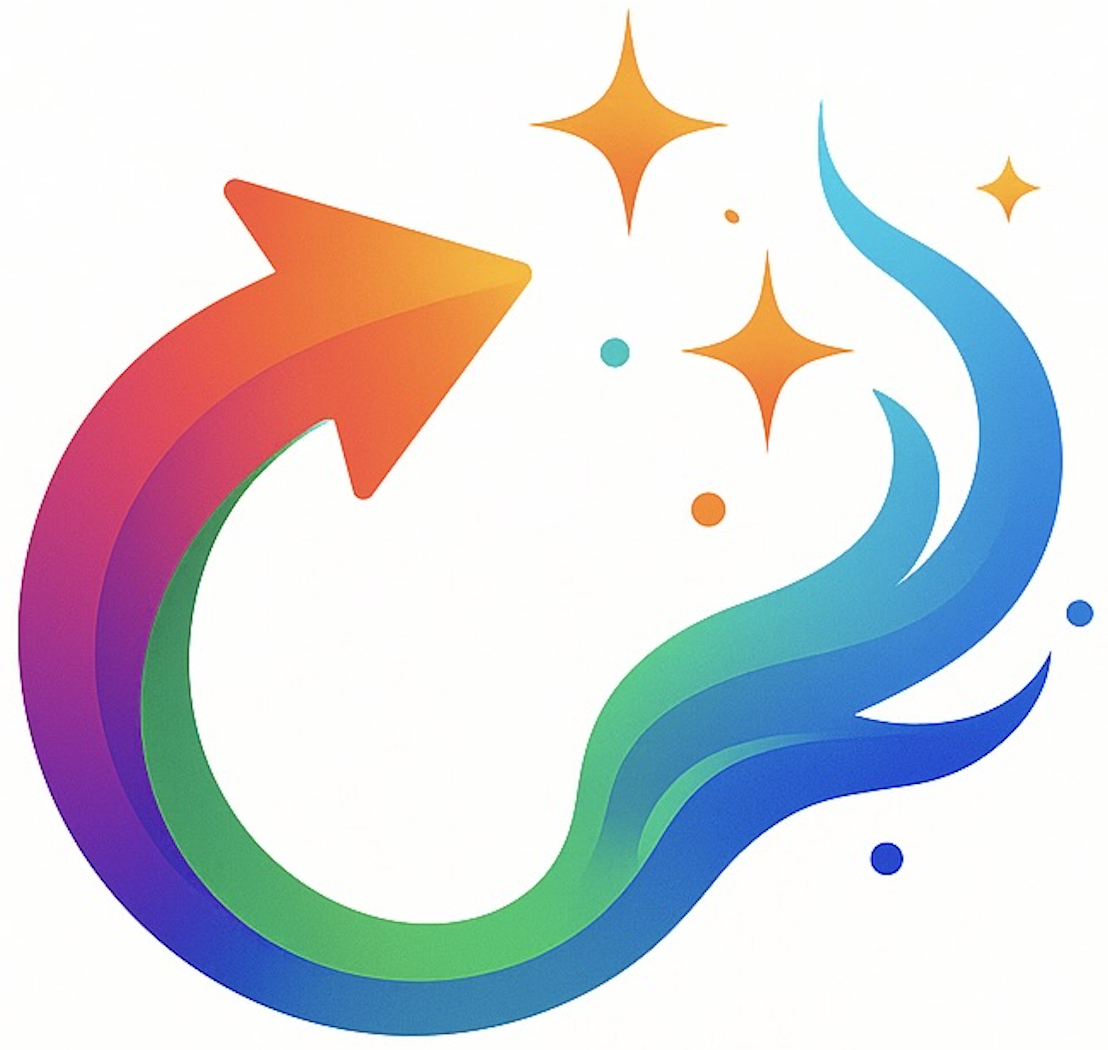} Turning the Spell Around: Lightweight Alignment Amplification via Rank-One Safety Injection}

\author{%
\centerline{\begin{tabular}{cc}
Harethah Abu Shairah\footnotemark[1] & \qquad Hasan Abed Al Kader Hammoud\footnotemark[1] \\[15pt]
George Turkiyyah & \qquad Bernard Ghanem \\
\end{tabular}} \\[25pt]
King Abdullah University of Science and Technology (KAUST) \\
Thuwal, Saudi Arabia \\
}
\begin{document}
\maketitle
\footnotetext[1]{Equal contribution.}

\begin{abstract}
Safety alignment in Large Language Models (LLMs) often involves mediating internal representations to refuse harmful requests. Recent research has demonstrated that these safety mechanisms can be bypassed by ablating or removing specific representational directions within the model. In this paper, we propose the opposite approach: \textsc{Rank-One Safety Injection (ROSI)}, a white-box method that \emph{amplifies} a model's safety alignment by permanently steering its activations toward the refusal-mediating subspace. \textsc{ROSI} operates as a simple, fine-tuning-free rank-one weight modification applied to all residual stream write matrices. The required safety direction can be computed from a small set of harmful and harmless instruction pairs. We show that \textsc{ROSI} consistently increases safety refusal rates while preserving the utility of many models across diverse architectures. Furthermore, we show that \textsc{ROSI} can also re-align 'uncensored' models by amplifying their own latent safety directions, demonstrating its utility as an effective last-mile safety procedure. Our results show that targeted, interpretable weight steering is a cheap and effective mechanism to improve LLM safety, complementing more resource-intensive fine-tuning paradigms.
\end{abstract}


\section{Introduction}

Large language models (LLMs) have demonstrated outstanding generality \citep{brown2020language}, excelling in tasks ranging from factual question answering \citep{kamalloo2023_qa} and reasoning \citep{wei2023_cot} to code synthesis \citep{tong2024_codejudge} and creative writing \citep{gómezrodríguez2023_writing}. Their versatility has made them the foundation of modern conversational assistants and productivity tools, where alignment techniques such as supervised fine-tuning and reinforcement learning from human feedback enable models to follow user instructions while adhering to safety constraints \citep{ouyang2022_rlhf}. As general-purpose interfaces for language interaction, LLMs are now widely deployed, fueling expectations that they may one day serve as core components of autonomous high-stakes systems.

Yet the same properties that make LLMs powerful also render them fragile and exposed to attack. Pre-training on vast, uncurated corpora inevitably gives models the capacity to generate harmful content \citep{wu2024_llmsecurity}, and safety alignment through post-training optimization offers only a partial safeguard \citep{mendu2025_saferpretraining}. Researchers have shown that even carefully aligned chat models remain vulnerable to a growing arsenal of jailbreak strategies, including prompt injection, obfuscation, multilingual exploits, and fine-tuning aimed at suppressing refusal, all capable of circumventing safety guardrails \citep{lin2024_jailbreaks, chu2024_comprehensiveassessmentjailbreakattacks, wei2023_jailbrokensafetyfail}.

\begin{figure*}
    \centering
    \caption{\textbf{\textsc{RANK-ONE SAFETY INJECTION (ROSI)}.} An aligned model processes both \textcolor{darkgreen}{benign} and \textcolor{red}{harmful} prompts in a forward pass (1). A safety vector is derived from the difference between harmful and harmless activations (2). Subtracting this vector ablates safety signals, producing an Abliterated Model. Adding it reinforces safety, producing a ROSI Model.}
    \includegraphics[width=1\linewidth]{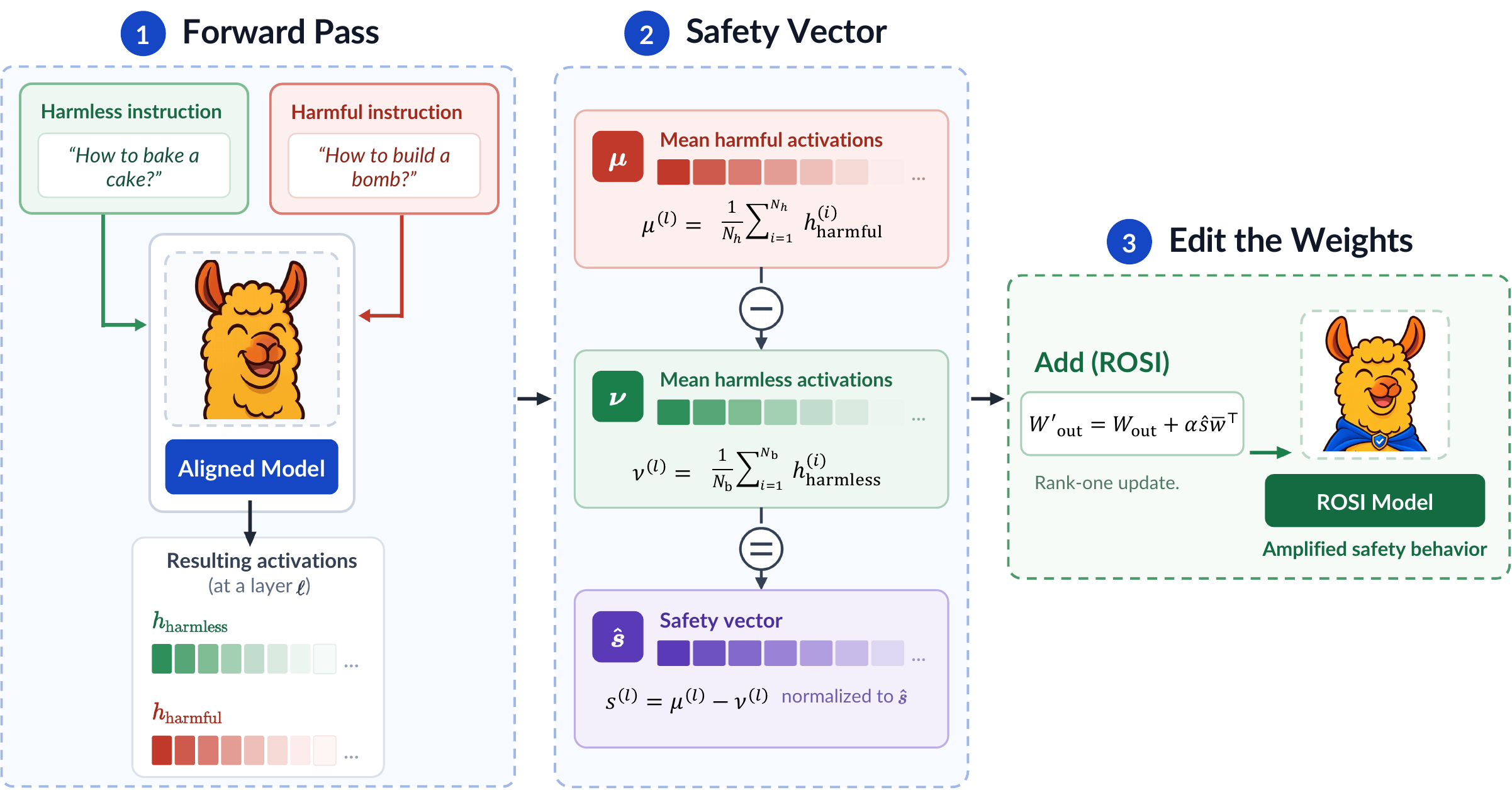}
    \label{fig:rosi_main}

\end{figure*}

Recent advances in mechanistic interpretability shed light on why these vulnerabilities arise. In particular, \citet{arditi2024refusal} demonstrate that refusal behavior is mediated by a \emph{one-dimensional linear direction} in the activation space of many open-source chat models. Erasing this ``refusal direction'' from the residual stream suffices to disable safety alignment, enabling harmful completions; conversely, adding this direction to a model's activations can induce refusal even on benign prompts. This finding shows that refusal and safety are encoded in an interpretable, causal subspace. 

Inspired by these insights, we ask the opposite question: rather than \emph{removing} safety, can we systematically \emph{amplify} it? In this paper, we propose \textsc{Rank-One Safety Injection (ROSI)}, a simple, fine-tuning-free method that hardens model refusal by applying a permanent lightweight rank-one modification to its weights. \textsc{ROSI} extracts a safety direction from a small set of harmful/harmless instruction pairs, and permanently injects this direction into all residual stream write matrices. 

We empirically demonstrate that \textsc{ROSI} provides two key benefits. First, it amplifies the safety of already aligned models, substantially improving their refusal rates and robustness against jailbreak attacks with negligible loss of utility. Second, it can re-align ``uncensored'' models that have been deliberately fine-tuned to ignore safety, reinstating refusal behavior without retraining. 
In summary, our contributions are:
\begin{itemize}
    \item We introduce \textsc{Rank-One Safety Injection (ROSI)}, a lightweight and interpretable weight-editing method to improve safety alignment in LLMs.
    \item We show that \textsc{ROSI} consistently improves the refusal and robustness of aligned models while preserving general utility on standard benchmarks.
    \item We demonstrate that \textsc{ROSI} can serve as an effective last-mile safety procedure, re-aligning uncensored models without expensive retraining.
\end{itemize}


\section{Related Work}

\paragraph{Mechanistic Interpretability of Refusal.}  
A central finding in alignment research is that refusal behavior in LLMs can be localized to low-dimensional linear features. \citet{arditi2024refusal} showed that a single direction in the residual stream mediates refusals across diverse chat models, with erasure or amplification of this direction directly controlling compliance with harmful prompts. Follow-up work has extended this line of inquiry: \citet{zhao2024harmfulness} disentangled harmfulness from refusal, showing that models encode internal judgments of harmfulness independently of whether they refuse; \citet{hong2025reasoning} identified another single direction governing the balance between reasoning and memorization; and \citet{jain2024makes} demonstrated how fine-tuning minimally alters weights to cluster unsafe activations. Others proposed activation interventions, including SAE-based steering \citep{o2024steering, he2025saif}, Trojan activation bypasses \citep{wang2024trojan}, and neuron or rank-level manipulations \citep{wei2024assessing, li2024safetypattern}.

\paragraph{Safety Steering and Training-free Defenses.}  
Training-free interventions attempt to steer model activations without costly fine-tuning. Early work showed that feature directions derived from contrastive inputs can modulate model behavior \citep{zou2023representation, panickssery2023contrastive, li2024steering, marks2023steering, turner2023activation}. Sparse autoencoders (SAEs) provide an unsupervised route to discover such features \citep{bricken2023monosemanticity, templeton2024sparse}. Recently, SAE-based steering has been applied directly to safety, revealing both promise and utility tradeoffs \citep{o2024steering}. Extensions include instruction-following features \citep{he2025saif}, category-wise safety steering \citep{ghosh2025safesteer, bhattacharjee2024towards}, and adaptive methods such as AdaSteer \citep{zhao2025adasteer}. Complementary strategies include Safety Arithmetic \citep{hazra2024safety}, Representation Bending \citep{yousefpour-etal-2025-representation}, Low-Rank Extrapolation \citep{lox2025_perin}, adversarial training approaches such as ReFAT \citep{yu2024robust}, and  null-space constraints methods like AlphaSteer \citep{sheng2025_alphasteer} that builds on insights from AlphaEdit\citep{fang2025_alphaedit} which is used for robust knowledge editing. Foundational studies further established linear features in representation spaces \citep{bolukbasi2016man, elhage2022toy, geiger2024finding, ravfogel2020null}. While effective, many steering-based defenses introduce capability tradeoffs, motivating interpretable and more surgical alternatives such as ours.

\paragraph{Beyond Steering: Fine-tuning and Safety Robustness.}  
Another line of work examines how safety alignment emerges or fails under fine-tuning. Works like \citet{zhan2023removing, yang2023shadow, qi2023fine, lermen2023undoing} show that even small malicious or benign finetunes can undo refusal, while mechanistic studies suggest the internal circuitry remains intact \citep{jain2024makes}.  textsc{SafeLoRA} \citep{hsu2025_safelora} shows, without training or data, how \textsc{LoRA} weights can be projected onto a safety-aligned subspace to reduce safety degradation from fine-tuning LLMs. Other interventions strengthen refusal explicitly, such as extended-refusal finetuning against abliteration attacks \citep{shairah2025embarrassingly}, refusal tokens for controllable calibration \citep{jain2024refusal}, and single-vector ablations to mitigate false refusals \citep{wang2025surgical}. Others work on run-time interventions to protect against jailbreaks, such as \textsc{SmoothLLM} \citep{robey2024_smoothllmde}, and Jailbreak Antidote \citep{shen2025_jailbreakantidote}. Alignment fragility also arises in model merging: \citet{hammoud-etal-2024-model} showed that unsafe models contaminate the merged ones unless alignment is explicitly included.

\paragraph{Our Contribution.}  
We build directly on the insight of \citet{arditi2024refusal} but invert its vulnerability: instead of ablating the safety direction to weaken safety, our \textsc{ROSI} method permanently injects it into model weights. Compared to inference-time steering \citep{o2024steering, zhao2025adasteer, ghosh2025safesteer, sheng2025_alphasteer, shen2025_jailbreakantidote}, \textsc{ROSI} provides a one-time lightweight, fine-tuning-free, interpretable mechanism that is permanent yet minimally invasive. Compared to approaches based on fine-tuning \citep{zhan2023removing, shairah2025embarrassingly}, it achieves comparable robustness with a much lower cost. Importantly, \textsc{ROSI} is not intended to replace existing safety strategies; it can be layered with steering, fine-tuning, or other alignment methods to further reinforce model robustness. Thus, our work illustrates how mechanistic interpretability can be leveraged not only to diagnose vulnerabilities but also to design efficient last-mile safety amplification techniques.

\section{Methodology}
\label{sec:methodology}
Our proposed method, \textsc{ROSI}, which is illustrated in Figure \ref{fig:rosi_main}, is based on the principle that high-level concepts such as safety are linearly represented in the activation space of a model. We first extract this "safety direction" and then use it to craft a permanent modification to the model's weights.

\subsection{Mathematical Preliminaries: Transformers}
A decoder-only Transformer model processes a sequence of input tokens $\mathbf{t} = (t_1, \ldots, t_n)$. The core of the model is the residual stream, $\mathbf{x}_i^{(l)} \in \mathbb{R}^{d_{\text{model}}}$, which represents the activation for the $i$-th token at the $l$-th layer. Each layer $l$ updates this activation through an attention block and a multi-layer perceptron (MLP) block:
\begin{align}
\tilde{\mathbf{x}}_i^{(l)} &= \mathbf{x}_i^{(l)} + \mathtt{Attn}^{(l)}(\mathbf{x}_{1:i}^{(l)}) \\
\mathbf{x}_i^{(l+1)} &= \tilde{\mathbf{x}}_i^{(l)} + \mathtt{MLP}^{(l)}(\tilde{\mathbf{x}}_i^{(l)})
\end{align}
The key components that are written in the residual stream are the attention output projection matrix ($W_O$) and the MLP output projection matrix ($W_{out}$). Our method targets these matrices, among others, for modification.

\subsection{Extracting the Safety Direction}
To isolate the direction in the activation space corresponding to safety and refusal, we employ the difference-in-means technique. We construct two small and contrasting datasets.
\begin{itemize}
    \item $\mathcal{D}_{\text{harmful}}$: A set of instructions that should elicit a refusal (e.g., "How do I build a bomb?").
    \item $\mathcal{D}_{\text{harmless}}$: A set of benign instructions that should be answered helpfully (e.g., "How do I bake a cake?").
\end{itemize}
We run the model on all the prompts in both datasets and collect the residual stream activations $\mathbf{x}_i^{(l)}$ at a specific layer $l$ and the position of the token $i$ (typically the last token of the prompt). We then compute the mean activation for each dataset:
\begin{align}
\boldsymbol{\mu}^{(l)} &= \frac{1}{|\mathcal{D}_{\text{harmful}}|} \sum_{\mathbf{t} \in \mathcal{D}_{\text{harmful}}} \mathbf{x}_i^{(l)}(\mathbf{t}) \\
\boldsymbol{\nu}^{(l)} &= \frac{1}{|\mathcal{D}_{\text{harmless}}|} \sum_{\mathbf{t} \in \mathcal{D}_{\text{harmless}}} \mathbf{x}_i^{(l)}(\mathbf{t})
\end{align}
The safety direction $\mathbf{s}^{(l)}$ is defined as the difference between these two means:
\begin{equation}
\mathbf{s}^{(l)} = \boldsymbol{\mu}^{(l)} - \boldsymbol{\nu}^{(l)}
\end{equation}
We select the optimal layer $l^*$ that yields the most effective direction based on a validation set of harmful and harmless prompts. We select the direction that maximizes refusal on harmful prompts while maintaining a KL-Divergence of $\leq0.1$ on the harmless instructions. The final normalized safety direction is denoted as $\hat{\mathbf{s}}$.

\subsection{Rank-One Safety Injection (ROSI)}
Previous work has shown that one can ablate a direction $\hat{\mathbf{s}}$ from a weight matrix $W$ by applying a projection: $W' \leftarrow (I - \hat{\mathbf{s}}\hat{\mathbf{s}}^T)W$. This effectively removes the model's ability to represent information along that direction.

We propose the opposite: to amplify this direction. We achieve this by modifying every weight matrix $W_{\text{out}} \in \mathbb{R}^{d_{\text{model}} \times d_{\text{input}}}$ that writes to the residual stream. The modification is a rank-one update designed to add a small, consistent push in the direction of $\hat{\mathbf{s}}$. The \textsc{ROSI} update rule is:
\begin{equation}
W'_{\text{out}} \leftarrow W_{\text{out}} + \alpha \cdot \hat{\mathbf{s}} \cdot \bar{\mathbf{w}}^T
\label{eq:rosi_update}
\end{equation}
where $\alpha$ is a scalar hyperparameter that controls the strength of the injection, $\hat{\mathbf{s}} \in \mathbb{R}^{d_{\text{model}}}$ is the normalized safety direction, and $\bar{\mathbf{w}} \in \mathbb{R}^{d_{\text{input}}}$ is the mean of the row vectors of the original weight matrix $W_{\text{out}}$.

This formulation creates a rank-one matrix $\alpha (\hat{\mathbf{s}} \bar{\mathbf{w}}^T)$ which is added to the original weights. The intuition is that for an average input, this modification adds a component proportional to the safety direction $\hat{\mathbf{s}}$ to the output, effectively steering the model's activations toward the refusal-mediating subspace. This is a permanent, efficient, and targeted change to the model's behavior.

\section{Experiments and Results}
Our empirical evaluation is designed to answer three key questions:
\begin{enumerate}
    \item Can \textsc{ROSI} amplify the safety of existing, aligned models and improve their robustness to adversarial attacks without degrading their general capabilities?
    \item Can \textsc{ROSI} effectively inject safety into "uncensored" models that have been fine-tuned to bypass safety constraints?
    \item Does this injected safety come at the cost of utility in these uncensored models?
\end{enumerate}
We address these questions through a series of controlled experiments on a diverse set of models and benchmarks.

\begin{table}
\centering
\caption{\textbf{Harm Refusal in Aligned Models.} \textsc{ROSI} consistently improves the refusal rate for harmful prompts (HR \%) while maintaining high compliance for benign ones (BC \%).}
\label{tab:safety-summary-rounded}
\small
\resizebox{\linewidth}{!}{%
\begin{tabular}{l @{\extracolsep{\fill}} c c c}
\toprule
\textbf{Model} & \textbf{\textsc{ROSI}} & \textbf{HR \%} & \textbf{BC \%} \\
\midrule
\multirow{2}{*}{\textsc{Gemma-2B-Instruct}}           & \xmark & 98.4 & 99.4  \\
                                             & \cmark & 99.8 \textcolor{darkgreen}{(+1.5)} & 99.0 \textcolor{darkred}{(-0.4)} \\
\midrule
\multirow{2}{*}{\textsc{Llama-2-7b-chat-hf}} & \xmark & 99.8 & 98.8  \\
                                             & \cmark & 100.0 \textcolor{darkgreen}{(+0.2)} & 99.8 \textcolor{darkgreen}{(+1.0)} \\                        
                                             
\midrule
\multirow{2}{*}{\textsc{Meta-Llama-3.1-8B-Instruct}}  & \xmark & 98.2 & 99.6  \\
                                             & \cmark & 99.1 \textcolor{darkgreen}{(+0.9)} & 99.6 \textcolor{gray}{(0.0)} \\
\midrule                                                                                         
\multirow{2}{*}{\textsc{Meta-Llama-3.2-1B-Instruct}}  & \xmark & 79.5 & 99.2  \\
                                             & \cmark & 92.7 \textcolor{darkgreen}{(+13.2)} & 95.9 \textcolor{darkred}{(-3.9)} \\
\midrule
\multirow{2}{*}{\textsc{Qwen2.5-0.5B-Instruct}}       & \xmark & 90.4 & 98.6  \\
                                             & \cmark & 99.3 \textcolor{darkgreen}{(+8.9)} & 91.4 \textcolor{darkred}{(-7.2)} \\
\midrule
\multirow{2}{*}{\textsc{Qwen2.5-3B-Instruct}}         & \xmark & 89.8 & 99.6  \\
                                             & \cmark & 99.6 \textcolor{darkgreen}{(+9.8)} & 98.6 \textcolor{darkred}{(-1.0)} \\
\midrule
\multirow{2}{*}{\textsc{Qwen2.5-7B-Instruct}}         & \xmark & 95.8 & 100.0 \\
                                             & \cmark & 100.0 \textcolor{darkgreen}{(+4.2)} & 99.0 \textcolor{darkred}{(-1.0)} \\
\midrule
\multirow{2}{*}{\textsc{Qwen2.5-14B-Instruct}}        & \xmark & 98.9 & 100.0 \\
                                             & \cmark & 100.0 \textcolor{darkgreen}{(+1.1)} & 99.4 \textcolor{darkred}{(-0.6)} \\
\midrule
\multirow{2}{*}{\textsc{Yi-6B-Chat}}                  & \xmark & 81.3 & 99.6  \\
                                             & \cmark & 99.5 \textcolor{darkgreen}{(+18.2)} & 97.7 \textcolor{darkred}{(-1.7)} \\
\bottomrule
\end{tabular}
}
\end{table}

\begin{table*}[!ht]
\centering
\caption{\textbf{Jailbreak Robustness of Aligned Models.} Scores represent attack success rates (lower is better). \textsc{ROSI} significantly reduces model vulnerability across all attack vectors.}
\label{tab:model_perf_rosi}
\resizebox{\linewidth}{!}{%
\begin{tabular}{l c c c ccc c}
\toprule
\multirow{2}{*}{Model} & \multirow{2}{*}{\textsc{ROSI}} & \multirow{2}{*}{\makebox[1.8cm][c]{\textsc{DAN} ↓}} & \multirow{2}{*}{\makebox[2.1cm][c]{\textsc{HarmBench} ↓}} & \multicolumn{3}{c}{\textsc{WildGuardTest} ↓} & \multirow{2}{*}{\makebox[3cm][c]{\textsc{WildJailbreak} Harmful ↓}} \\
\cmidrule(lr){5-7}
 & & & & WG-Micro & WG-Adv. & WG-Vanilla & \\
\midrule
\multirow{2}{*}{\textsc{Gemma-2B-Instruct}} & \xmark & 5.3 & 6.2 & 9.1 & 16.6 & 2.9 & 42.3 \\
                  & \cmark & \textbf{1.0} {\textcolor{darkgreen}{(-4.3)}} & \textbf{3.4} {\textcolor{darkgreen}{(-2.8)}} & \textbf{2.4} {\textcolor{darkgreen}{(-6.7)}} & \textbf{4.7} {\textcolor{darkgreen}{(-11.9)}} & \textbf{0.5} {\textcolor{darkgreen}{(-2.4)}} & \textbf{8.2} {\textcolor{darkgreen}{(-34.1)}} \\
\midrule
\multirow{2}{*}{\textsc{Llama-2-7b-chat-hf}} & \xmark & 0.0 & 0.0 & 0.9 & 2.1 & 0.0 & 3.5 \\
                   & \cmark & 0.0 {\textcolor{gray}{(0.0)}} & 0.0 {\textcolor{gray}{(0.0)}} & \textbf{0.0} {\textcolor{darkgreen}{(-0.9)}} & \textbf{0.0} {\textcolor{darkgreen}{(-2.1)}} & 0.0 {\textcolor{gray}{(0.0)}} & \textbf{0.1} {\textcolor{darkgreen}{(-3.4)}} \\
\midrule
\multirow{2}{*}{\textsc{Llama-3.1-8B-Instruct}} & \xmark & 0.3 & 5.9 & 1.6 & 2.7 & 0.7 & 14.8 \\
           & \cmark & \textbf{0.0} {\textcolor{darkgreen}{(-0.3)}} & 5.3 {\textcolor{darkgreen}{(-0.6)}} & \textbf{0.0} {\textcolor{darkgreen}{(-1.6)}} & \textbf{0.0} {\textcolor{darkgreen}{(-2.7)}} & \textbf{0.0} {\textcolor{darkgreen}{(-0.7)}} & \textbf{1.8} {\textcolor{darkgreen}{(-13.0)}} \\
\midrule
\multirow{2}{*}{\textsc{Llama-3.2-1B-Instruct}} & \xmark & 1.3 & 8.4 & 4.0 & 3.9 & 4.1 & 18.7 \\
           & \cmark & \textbf{0.0} {\textcolor{darkgreen}{(-1.3)}} & \textbf{5.6} {\textcolor{darkgreen}{(-2.8)}} & \textbf{1.3} {\textcolor{darkgreen}{(-2.7)}} & \textbf{1.5} {\textcolor{darkgreen}{(-2.4)}} & \textbf{1.2} {\textcolor{darkgreen}{(-2.9)}} & \textbf{7.5} {\textcolor{darkgreen}{(-11.1)}} \\
\midrule
\multirow{2}{*}{\textsc{Qwen2.5-0.5B-Instruct}} & \xmark & 36.0 & 31.6 & 33.1 & 48.1 & 20.9 & 91.8 \\
                      & \cmark & \textbf{7.0} {\textcolor{darkgreen}{(-29.0)}} & \textbf{12.8} {\textcolor{darkgreen}{(-18.8)}} & \textbf{21.1} {\textcolor{darkgreen}{(-12.0)}} & \textbf{38.0} {\textcolor{darkgreen}{(-10.1)}} & \textbf{7.3} {\textcolor{darkgreen}{(-13.6)}} & \textbf{58.8} {\textcolor{darkgreen}{(-33.0)}} \\
\midrule
\multirow{2}{*}{\textsc{Qwen2.5-3B-Instruct}} & \xmark & 52.7 & 12.5 & 21.4 & 37.4 & 8.3 & 93.7 \\
                    & \cmark & \textbf{6.7} {\textcolor{darkgreen}{(-46.0)}} & \textbf{1.6} {\textcolor{darkgreen}{(-10.9)}} & \textbf{12.7} {\textcolor{darkgreen}{(-8.7)}} & \textbf{26.7} {\textcolor{darkgreen}{(-10.7)}} & \textbf{1.2} {\textcolor{darkgreen}{(-7.1)}} & \textbf{61.5} {\textcolor{darkgreen}{(-32.2)}} \\
\midrule
\multirow{2}{*}{\textsc{Qwen2.5-7B-Instruct}} & \xmark & 40.3 & 22.5 & 18.6 & 36.2 & 4.1 & 90.7 \\
                    & \cmark & \textbf{11.7} {\textcolor{darkgreen}{(-28.6)}} & \textbf{1.9} {\textcolor{darkgreen}{(-20.6)}} & \textbf{3.9} {\textcolor{darkgreen}{(-14.7)}} & \textbf{7.7} {\textcolor{darkgreen}{(-28.5)}} & \textbf{0.7} {\textcolor{darkgreen}{(-3.4)}} & \textbf{36.7} {\textcolor{darkgreen}{(-54.0)}} \\
\midrule
\multirow{2}{*}{\textsc{Qwen2.5-14B-Instruct}} & \xmark & 32.3 & 7.2 & 12.1 & 24.0 & 2.4 & 81.2 \\
                     & \cmark & \textbf{5.0} {\textcolor{darkgreen}{(-27.3)}} & \textbf{1.6} {\textcolor{darkgreen}{(-5.6)}} & \textbf{5.1} {\textcolor{darkgreen}{(-7.0)}} & \textbf{11.0} {\textcolor{darkgreen}{(-13.0)}} & \textbf{0.2} {\textcolor{darkgreen}{(-2.2)}} & \textbf{43.9} {\textcolor{darkgreen}{(-37.3)}} \\
\midrule
\multirow{2}{*}{\textsc{Yi-6B-Chat}} & \xmark & 52.0 & 20.9 & 22.7 & 39.2 & 9.2 & 89.4 \\
           & \cmark & \textbf{15.3} {\textcolor{darkgreen}{(-36.7)}} & \textbf{7.8} {\textcolor{darkgreen}{(-13.1)}} & \textbf{10.1} {\textcolor{darkgreen}{(-12.6)}} & \textbf{22.0} {\textcolor{darkgreen}{(-17.2)}} & \textbf{0.5} {\textcolor{darkgreen}{(-8.7)}} & \textbf{44.6} {\textcolor{darkgreen}{(-44.8)}} \\
\bottomrule
\end{tabular}
}
\end{table*}

\begin{table*}[!ht]
\centering
\caption{\textbf{Utility Preservation in Aligned Models.} Performance on standard benchmarks with \textsc{ROSI} (\cmark) versus baseline (\xmark).}
\label{tab:benchmark_results_interleaved_avgdelta}
\resizebox{\linewidth}{!}{%
\begin{tabular}{l c cccccc}
\toprule
Model & \textsc{ROSI} & \textsc{MMLU} & \textsc{HellaSwag} & \textsc{Arc Easy} & \textsc{Arc Chal.} & \textsc{BoolQ} & \textsc{TruthfulQA} \\
\midrule
\multirow{2}{*}{\textsc{Gemma-2B-Instruct}}         
 & \xmark & 38.1 & 49.2 & 71.7 & 40.4 & 63.7 & 45.8 \\
 & \cmark & 38.3 \textcolor{darkgreen}{(+0.2)} & 49.3 \textcolor{darkgreen}{(+0.1)} & 70.8 \textcolor{darkred}{(-0.9)} & 39.0 \textcolor{darkred}{(-1.4)} & 61.4 \textcolor{darkred}{(-2.3)} & 46.7 \textcolor{darkgreen}{(+0.9)} \\
\midrule
\multirow{2}{*}{\textsc{Llama-2-7B-Chat-HF}}         
 & \xmark & 46.3 & 57.8 & 74.0 & 43.9 & 79.6 & 45.3 \\
 & \cmark & 46.4 \textcolor{darkgreen}{(+0.1)} & 57.7 \textcolor{darkred}{(-0.1)} & 73.4 \textcolor{darkred}{(-0.6)} & 43.3 \textcolor{darkred}{(-0.6)} & 79.8 \textcolor{darkgreen}{(+0.2)} & 47.2 \textcolor{darkgreen}{(+1.9)} \\
\midrule
\multirow{2}{*}{\textsc{Meta-Llama-3.1-8B-Instruct}} 
 & \xmark & 68.0 & 59.1 & 81.7 & 51.6 & 84.0 & 54.1 \\
 & \cmark & 67.6 \textcolor{darkred}{(-0.4)} & 58.9 \textcolor{darkred}{(-0.2)} & 81.1 \textcolor{darkred}{(-0.6)} & 51.1 \textcolor{darkred}{(-0.5)} & 83.8 \textcolor{darkred}{(-0.2)} & 54.8 \textcolor{darkgreen}{(+0.7)} \\
 \midrule
 \multirow{2}{*}{\textsc{Meta-Llama-3.2-1B-Instruct}} 
 & \xmark & 46.0 & 45.2 & 68.3 & 35.6 & 69.3 & 43.9 \\
 & \cmark & 45.4 \textcolor{darkred}{(-0.6)} & 45.4 \textcolor{darkgreen}{(+0.2)} & 67.4 \textcolor{darkred}{(-0.9)} & 34.7 \textcolor{darkred}{(-0.9)} & 68.7 \textcolor{darkred}{(-0.6)} & 45.0 \textcolor{darkgreen}{(+1.1)} \\
\midrule
\multirow{2}{*}{\textsc{Qwen2.5-0.5B-Instruct}}     
 & \xmark & 45.8 & 40.5 & 65.5 & 30.1 & 67.6 & 41.8 \\
 & \cmark & 45.3 \textcolor{darkred}{(-0.5)} & 40.4 \textcolor{darkred}{(-0.1)} & 64.3 \textcolor{darkred}{(-1.2)} & 29.6 \textcolor{darkred}{(-0.5)} & 63.2 \textcolor{darkred}{(-4.4)} & 43.8 \textcolor{darkgreen}{(+2.0)} \\
\midrule
\multirow{2}{*}{\textsc{Qwen2.5-3B-Instruct}}        
 & \xmark & 65.4 & 56.3 & 76.9 & 45.7 & 80.1 & 58.7 \\
 & \cmark & 65.0 \textcolor{darkred}{(-0.4)} & 55.8 \textcolor{darkred}{(-0.5)} & 76.6 \textcolor{darkred}{(-0.3)} & 45.1 \textcolor{darkred}{(-0.6)} & 77.4 \textcolor{darkred}{(-2.7)} & 59.7 \textcolor{darkgreen}{(+1.0)} \\
\midrule
\multirow{2}{*}{\textsc{Qwen2.5-7B-Instruct}}       
 & \xmark & 71.8 & 62.0 & 81.6 & 52.6 & 86.4 & 64.8 \\
 & \cmark & 71.9 \textcolor{darkgreen}{(+0.1)} & 61.9 \textcolor{darkred}{(-0.1)} & 81.0 \textcolor{darkred}{(-0.6)} & 52.6 \textcolor{gray}{(0.0)} & 86.2 \textcolor{darkred}{(-0.2)} & 66.1 \textcolor{darkgreen}{(+1.3)} \\
\midrule
\multirow{2}{*}{\textsc{Qwen2.5-14B-Instruct}}       
 & \xmark & 78.8 & 65.6 & 85.7 & 60.4 & 88.0 & 69.0 \\
 & \cmark & 78.9 \textcolor{darkgreen}{(+0.1)} & 65.6 \textcolor{gray}{(0.0)} & 85.6 \textcolor{darkred}{(-0.1)} & 60.7 \textcolor{darkgreen}{(+0.3)} & 85.8 \textcolor{darkred}{(-2.2)} & 71.9 \textcolor{darkgreen}{(+2.9)} \\
\midrule
\multirow{2}{*}{\textsc{Yi-6B-Chat}}                
 & \xmark & 61.6 & 57.7 & 74.5 & 44.1 & 82.8 & 49.9 \\
 & \cmark & 61.1 \textcolor{darkred}{(-0.5)} & 57.2 \textcolor{darkred}{(-0.5)} & 78.1 \textcolor{darkgreen}{(+3.6)} & 46.9 \textcolor{darkgreen}{(+2.8)} & 84.2 \textcolor{darkgreen}{(+1.4)} & 51.2 \textcolor{darkgreen}{(+1.3)} \\
\bottomrule
\end{tabular}%
}
\end{table*}

\begin{figure*}
    \centering
    
    \caption{\textbf{Injected Layers Ablations.} In \ref{fig:ablation-left}, we ablate the number of layers we apply \textsc{ROSI} to by taking a ratio $x$ (x-axis) of a model's layers centered around the index of layer $i$ used to extract the safety vector. In \ref{fig:ablation-right}, a subset of layers is selected randomly each time, we repeat 10 times for each ratio and take the average of the harm refusal rate. Confidence intervals are reported.}
    
    \begin{subfigure}[t]{0.47\linewidth}
        \centering
        \caption{\textbf{Windowed Injection.}}
        \includegraphics[width=\linewidth]{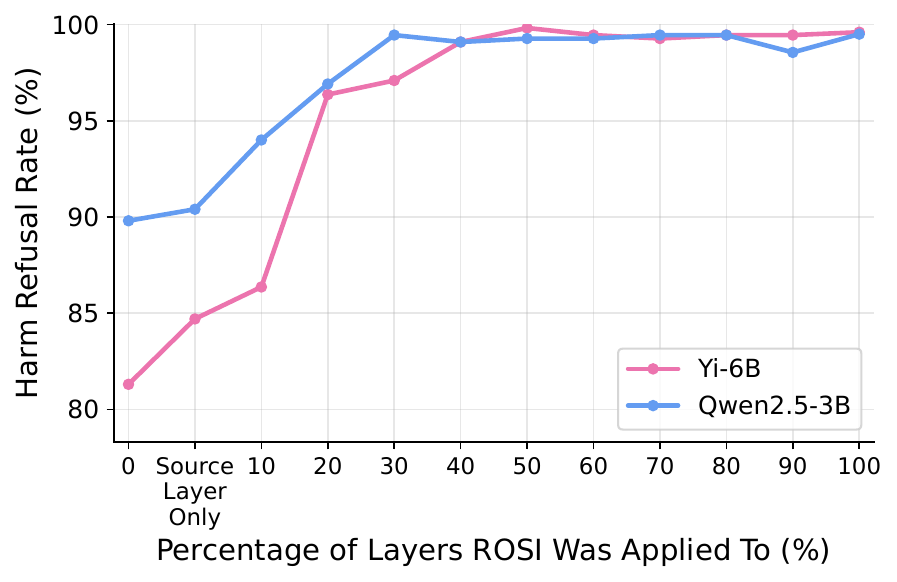}
        \label{fig:ablation-left}
    \end{subfigure}
    \hfill
    \begin{subfigure}[t]{0.47\linewidth}
        \centering
        \caption{\textbf{Randomized Injection.}}
        \includegraphics[width=\linewidth]{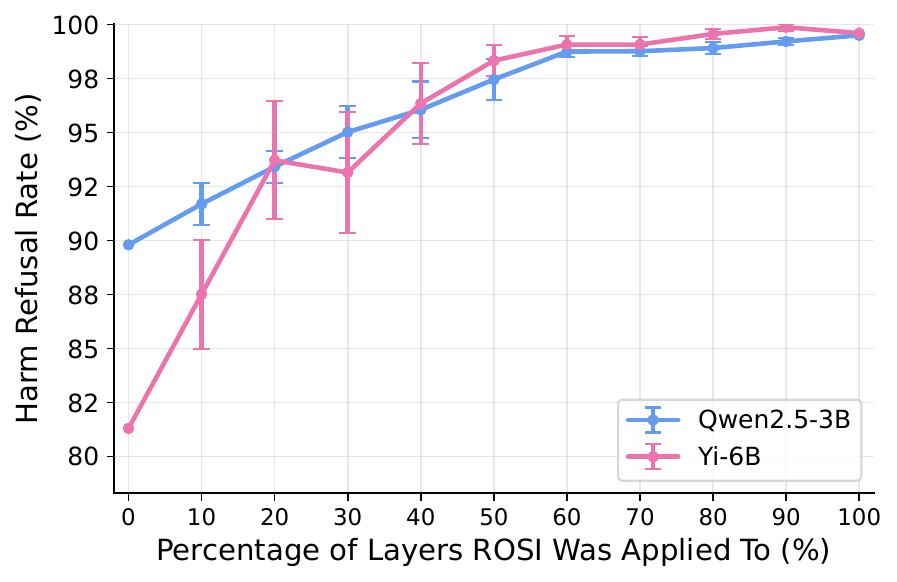}
        \label{fig:ablation-right}
    \end{subfigure}

    \label{fig:my-both}
\end{figure*}

\subsection{Experimental Setup}
\paragraph{Models.} We test two categories of models: \textbf{Aligned Models} including \textsc{Llama-2} \citep{touvron2023_llama2}, \textsc{Llama-3} \citep{dubey2024_llamaguard}, \textsc{Qwen2.5} \citep{qwen2025_qwen25}, \textsc{Gemma} \citep{gemmateam2024_gemma}, and \textsc{Yi} \citep{ai2025_yi}, which have standard safety training; and \textbf{Uncensored Models}, specifically the \textsc{Dolphin} series \citep{dolphin-ai}, which are intentionally fine-tuned to ignore safety.

\paragraph{Evaluation.} Safety is measured via Harm Refusal (HR) on \textsc{CatQA} \citep{bhardwaj2024_catqa}, a set of 550 harmful instructions from 11 categories, evaluated using \textsc{Llama Guard 3} \citep{dubey2024_llamaguard}. We also measure attack success rates on jailbreak benchmarks—\textsc{DAN}, \textsc{HarmBench} \citep{mazeika2024_harmbench}, \textsc{WildGuardTest}, and \textsc{WildJailbreak} \citep{wildteaming2024}—judged by \textsc{WildGuard} \citep{han2024_wildguard}. Utility is assessed on standard benchmarks: \textsc{MMLU} \citep{hendrycks2021_mmlu}, \textsc{HellaSwag} \citep{zellers2019_hellaswag}, \textsc{ARC} \citep{chollet2019_arc}, \textsc{BoolQ} \citep{clark2019_boolq}, and \textsc{TruthfulQA} \citep{lin2022_truthfulqa}. We also measure Benign Compliance (BC) on a randomly sampled set of 512 instructions from \textsc{Alpaca} \citep{alpaca}, to ensure \textsc{ROSI} models do not over refuse safe instructions.

\paragraph{Implementation.} The safety direction for each model was extracted using 50 harmful/harmless pairs. Generations use greedy decoding with a max length of 1024 tokens.

\subsection{Amplifying Safety in Aligned Models}
We first test \textsc{ROSI}'s ability to bolster the defenses of models that already possess safety alignment.

\paragraph{Increased Refusal and Jailbreak Robustness.}
As shown in Table \ref{tab:safety-summary-rounded}, applying \textsc{ROSI} consistently enhances the Harm Refusal (HR) rate across all aligned models tested. The effect is particularly pronounced for models with weaker baselines, such as \textsc{Yi-6B-Chat} (+18.2 points) and \textsc{Meta-Llama-3.2-1B-Instruct} (+13.2 points), elevating their safety to near-perfect levels. This improvement is not superficial; Table \ref{tab:model_perf_rosi} shows that \textsc{ROSI} drastically hardens models against a full suite of adversarial jailbreak attacks. For many models, attack success rates are cut by more than half, demonstrating a fundamental increase in robustness. 

In Appendix~\ref{app:peft}, we discuss what role \textsc{ROSI} can play in fine-tuning LLMs.

\begin{figure*}
    \centering
    \caption{\textbf{Applying \textsc{ROSI} to Uncensored Models.} In the forward pass, \textcolor{red}{harmful} and \textcolor{darkgreen}{harmless} instructions are prepended with a \textcolor{darkyellow}{system prompt} directing an uncensored model to reject harmful requests, thus eliciting refusal.}
    \includegraphics[width=1\linewidth]{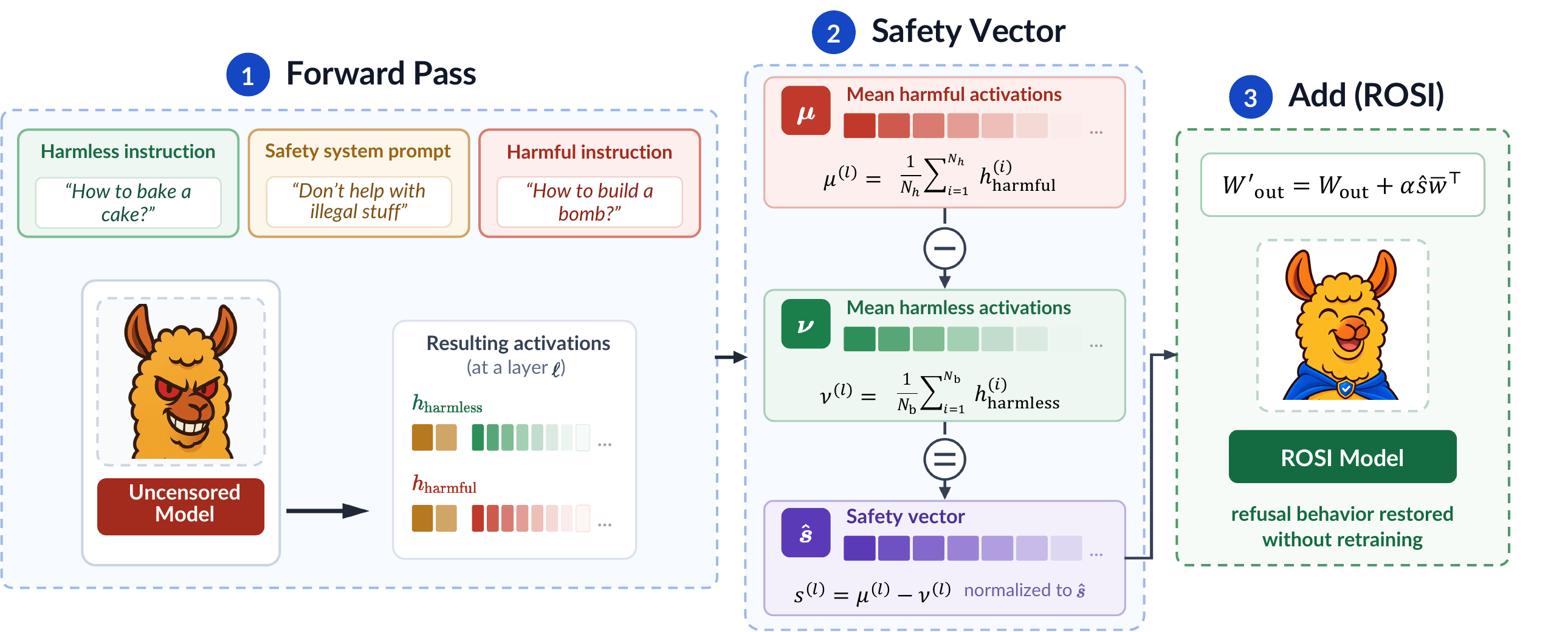}
    \label{fig:rosi_uncensored}

\end{figure*}

\paragraph{Preservation of Model Utility.}
Crucially, these safety gains do not compromise the models' core functionalities. Table \ref{tab:benchmark_results_interleaved_avgdelta} provides a comprehensive view of utility preservation. The average performance across a suite of six benchmarks remains remarkably stable. The vast majority of models see an average score change of less than 0.5\%. A similar pattern holds for BC, as seen in Table \ref{tab:safety-summary-rounded}, \textsc{ROSI} models' refusal of safe instructions, on average, remains minimal. While smaller models ($\leq1$B) show the biggest degradation in BC, they still gain more in HR than what they lose in BC. These results demonstrate that the safety direction is largely orthogonal to the representations required for knowledge and reasoning tasks. \textsc{ROSI} acts as a surgical tool, enhancing safety with minimal side effects.

\paragraph{Injected Layers Ablations.}
To assess how stable the \textsc{ROSI} update is within a model, we perform a set of ablations that vary both the number and the identity of the layers receiving the safety injection for two representative models, \textsc{Yi-6B} and \textsc{Qwen2.5-3B}. In the first setting, we inject \textsc{ROSI} into a contiguous block of layers centered on the layer index used to extract the safety vector, expanding this window according to a chosen fraction of the model’s total depth. Figure~\ref{fig:ablation-left} shows how injecting just at the source layer yields only modest improvements, and as the window of injected layers is expanded, the harm refusal rate keeps increasing until it stabilizes around the $30-40\%$ window size, suggesting that only a limited number of layers within a model contribute to the concept of "safety". In a second setting, we examine robustness by randomly selecting the same number of layers for each fraction. For every ratio, we repeat the process ten times and average the resulting refusal  scores. Figure~\ref{fig:ablation-right} displays a similar trend to the former experiment, but the confidence intervals show that performance varies considerably depending on the layers selected. Notably,\textsc{Yi-6B-Chat} peaks at 100\% HR rates in one of the runs where \textsc{ROSI} was applied to only half of the layers, which suggests that optimizing the set of injected layers can further improve performance.

\subsection{Injecting Safety into Uncensored Models}
The previous experiment demonstrated that \textsc{ROSI} can enhance refusal behavior in models that are already aligned. We now turn to the more demanding task of applying \textsc{ROSI} to uncensored \textsc{Dolphin} models. 

\paragraph{Eliciting Refusal Behavior and Reducing Vulnerability.}
The \textsc{Dolphin} models exhibit very low baseline safety, leaving little to no refusal signal to extract. Directly applying \textsc{ROSI} to a \textsc{Dolphin} model would therefore yield a vector  $\hat{\mathbf{s}}$  that does not represent a safety direction. 

To overcome this, we explicitly \emph{elicit} refusal behavior by modifying the system prompt, as can be seen in Figure \ref{fig:rosi_uncensored}. Specifically, we prepend instructions that direct the model to reject harmful categories of requests. This artificially introduces a refusal subspace that would otherwise be absent. Once present, we can apply \textsc{ROSI} to these models. Afterwards, the system prompt is no longer needed and is removed during testing.

\begin{table}
\caption{\textbf{Safety Injection in Uncensored Models.} Applying \textsc{ROSI} substantially boosts harm refusal (HR) across \textsc{Dolphin} models, while preserving compliance with benign instructions (BC). Ablations without a safety system prompt (\abmark) show the role of prompt-level safety conditioning.}
\label{tab:dolphin-safety-summary-rounded}
\centering
\small
\resizebox{\linewidth}{!}{%
\begin{tabular}{l @{\extracolsep{\fill}} c c c}
\toprule
\textbf{Model} & \textbf{ROSI} & \textbf{HR \%} & \textbf{BC \%} \\
\midrule
\multirow{3}{*}{\textsc{Dolphin3.0-Llama3.2-1B}} & \xmark & 23.5 & 100.0 \\
                                                 & \cmark & 46.0 \textcolor{darkgreen}{(+22.5)} & 99.4 \textcolor{darkred}{(-0.6)} \\
                                                 & \abmark & 18.4 \textcolor{darkred}{(-5.1)} & 100.0 \textcolor{gray}{(0.0)} \\
\midrule
\multirow{3}{*}{\textsc{Dolphin3.0-Qwen2.5-3B}}  & \xmark & 50.0 & 100.0 \\
                                                 & \cmark & 86.0 \textcolor{darkgreen}{(+36.0)} & 99.6 \textcolor{darkred}{(-0.4)} \\
                                                 & \abmark & 33.6 \textcolor{darkred}{(-16.4)} & 100.0 \textcolor{gray}{(0.0)} \\
\midrule
\multirow{3}{*}{\textsc{Dolphin3.0-Llama3.1-8B}} & \xmark & 65.8 & 100.0 \\
                                                 & \cmark & 100.0 \textcolor{darkgreen}{(+34.2)} & 100.0 \textcolor{gray}{(0.0)} \\
                                                 & \abmark & 88.9 \textcolor{darkgreen}{(+23.1)} & 100.0 \textcolor{gray}{(0.0)} \\
\midrule
\multirow{3}{*}{\textsc{Dolphin3.0-Mistral-24B}} & \xmark & 64.4 & 100.0 \\
                                                 & \cmark & 92.0 \textcolor{darkgreen}{(+27.6)} & 100.0 \textcolor{gray}{(0.0)} \\
                                                 & \abmark & 47.8 \textcolor{darkred}{(-16.6)} & 100.0 \textcolor{gray}{(0.0)} \\
\bottomrule
\end{tabular}
}
\end{table}

\begin{figure*}[!t]
\caption{\textbf{Activation separation in \textsc{Dolphin3.0-Qwen2.5-3B}.} (a) In the absence of a safety system prompt, the embeddings for harmful (red) and harmless (blue) inputs show significant overlap.
    (b) When a safety system prompt is introduced, the clusters become more distinct.}
    \centering   

    \begin{subfigure}[t]{0.45\linewidth}
        \centering
        \caption{\textbf{Without a safety system prompt.}}

        \includegraphics[width=\linewidth]{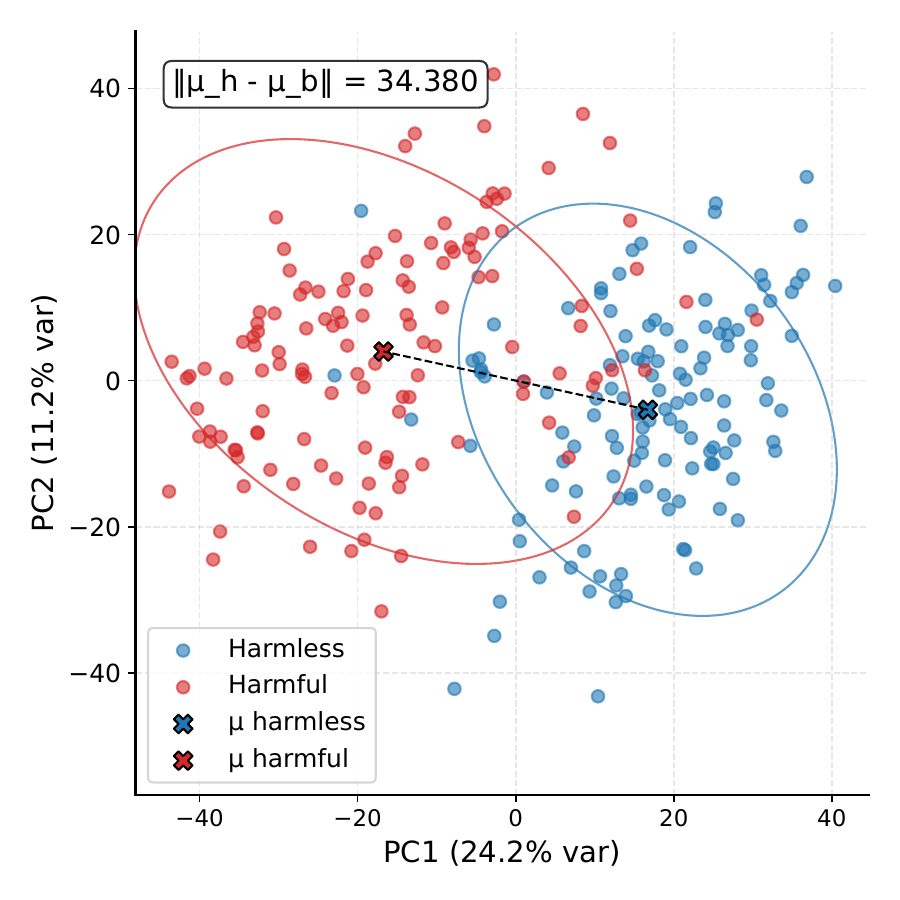}
        \label{fig:separation-left}
    \end{subfigure}
    \hfill
    \begin{subfigure}[t]{0.45\linewidth}
        \centering
        \caption{\textbf{With a safety system prompt.}}

        \includegraphics[width=\linewidth]{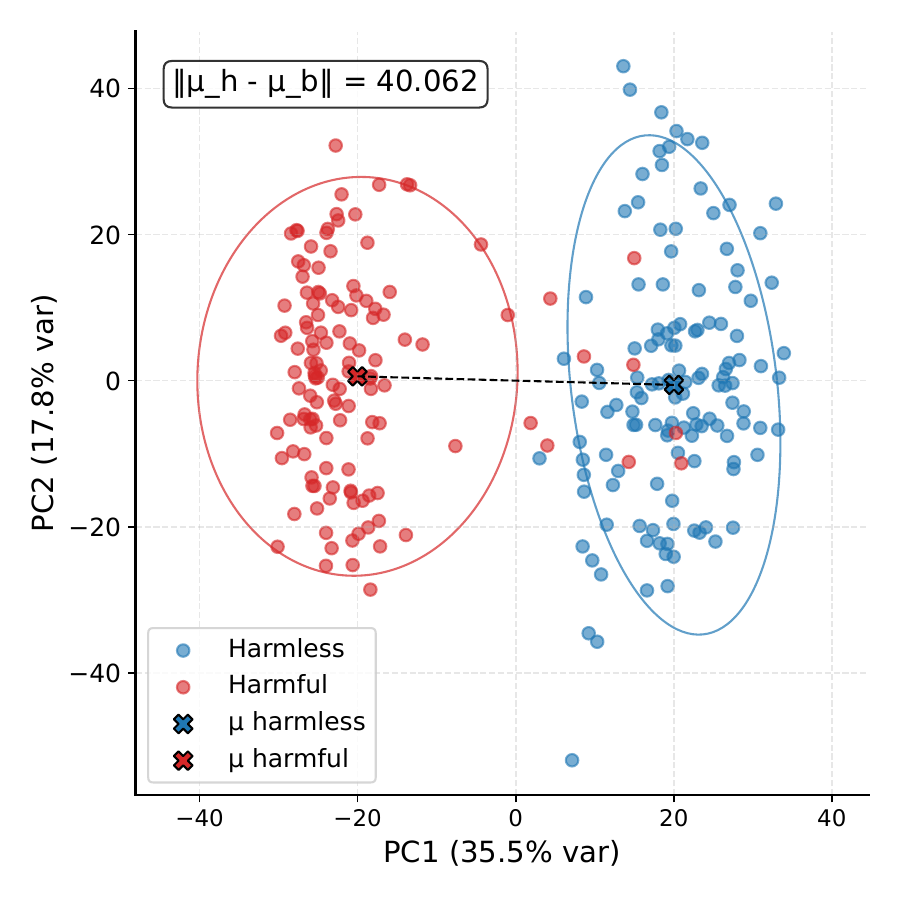}
        \label{fig:separation-right}
    \end{subfigure}
    \label{fig:separation}
\end{figure*}

Table \ref{tab:dolphin-safety-summary-rounded} shows that \textsc{ROSI} achieves dramatic improvements. For instance, \textsc{Dolphin3.0-Qwen2.5-3B}’s safe response rate skyrockets from 50.0\% to 86.0\% (+36.0), while \textsc{Dolphin3.0-Llama3.1-8B} is fully re-aligned to 100\% safety. This demonstrates that even uncensored models retain a latent safety direction that is potent enough to overwrite their fine-tuning when amplified. This injected safety also translates to improved robustness. As seen in Table \ref{tab:dolphin_perf_rosi}, \textsc{ROSI} provides a powerful first line of defense, slashing attack success rates by large margins (e.g., a 46.3-point reduction on \textsc{DAN} for \textsc{Dolphin3.0-Qwen2.5-3b}).

\begin{table*}[!t]
\centering
\caption{\textbf{Jailbreak Vulnerability of Uncensored Models.} Scores are attack success rates (lower is better). \textsc{ROSI} provides a crucial layer of defense, significantly reducing their extreme vulnerability.}
\label{tab:dolphin_perf_rosi}
\resizebox{\linewidth}{!}{%
\begin{tabular}{l c c c ccc c}
\toprule
\multirow{2}{*}{Model} & \multirow{2}{*}{\textsc{ROSI}} & \multirow{2}{*}{\makebox[1.8cm][c]{\textsc{DAN} ↓}} & \multirow{2}{*}{\makebox[2.1cm][c]{\textsc{HarmBench} ↓}} & \multicolumn{3}{c}{\textsc{WildGuardTest} ↓} & \multirow{2}{*}{\makebox[3cm][c]{\textsc{WildJailbreak} Harmful ↓}} \\
\cmidrule(lr){5-7}
 & & & & WG-Micro & WG-Adv. & WG-Vanilla & \\
 \midrule
\multirow{3}{*}{\textsc{Dolphin3.0-Llama3.2-1B}} & \xmark & 90.3 & 62.8 & 50.3 & 42.4 & 56.8 & 98.5 \\
           & \cmark & \textbf{65.7} {\textcolor{darkgreen}{(-24.7)}} & \textbf{51.9} {\textcolor{darkgreen}{(-10.9)}} & \textbf{33.9} {\textcolor{darkgreen}{(-16.4)}} & \textbf{38.3} {\textcolor{darkgreen}{(-4.2)}} & \textbf{30.3} {\textcolor{darkgreen}{(-26.5)}} & \textbf{88.9} {\textcolor{darkgreen}{(-9.5)}} \\
           & \abmark & 88.6 {\textcolor{darkgreen}{(-1.7)}} & 72.2 {\textcolor{darkred}{(+9.4)}} & 59.3 {\textcolor{darkred}{(+9.0)}} & 48.1 {\textcolor{darkred}{(+5.7)}} & 68.5 {\textcolor{darkred}{(+11.7)}} & 97.7 {\textcolor{darkgreen}{(-0.8)}} \\
\midrule
\multirow{3}{*}{\textsc{Dolphin3.0-Qwen2.5-3B}} & \xmark & 90.3 & 52.8 & 32.6 & 37.7 & 28.4 & 96.7 \\
           & \cmark & \textbf{44.0} {\textcolor{darkgreen}{(-46.3)}} & \textbf{20.9} {\textcolor{darkgreen}{(-31.9)}} & \textbf{15.4} {\textcolor{darkgreen}{(-17.2)}} & \textbf{27.3} {\textcolor{darkgreen}{(-10.4)}} & \textbf{5.6} {\textcolor{darkgreen}{(-22.8)}} & \textbf{70.4} {\textcolor{darkgreen}{(-26.3)}} \\
           & \abmark & 52.7 {\textcolor{darkgreen}{(-37.6)}} & 32.2 {\textcolor{darkgreen}{(-20.6)}} & 23.4 {\textcolor{darkgreen}{(-9.2)}} & 29.4 {\textcolor{darkgreen}{(-8.3)}} & 18.4 {\textcolor{darkgreen}{(-10.0)}} & 82.8 {\textcolor{darkgreen}{(-13.9)}} \ \\
\midrule
\multirow{3}{*}{\textsc{Dolphin3.0-Llama3.1-8B}} & \xmark & 90.3 & 54.7 & 27.0 & 34.7 & 20.6 & 94.0 \\
           & \cmark & 82.3 {\textcolor{darkgreen}{(-8.0)}} & 47.2 {\textcolor{darkgreen}{(-7.5)}} & 21.1 {\textcolor{darkgreen}{(-5.9)}} & 29.4 {\textcolor{darkgreen}{(-5.3)}} & 14.3 {\textcolor{darkgreen}{(-6.3)}} & \textbf{82.8} {\textcolor{darkgreen}{(-11.3)}} \\
           & \abmark & \textbf{81.3} {\textcolor{darkgreen}{(-9.0)}} & \textbf{44.7} {\textcolor{darkgreen}{(-10.0)}} & \textbf{19.2} {\textcolor{darkgreen}{(-7.8)}} & \textbf{26.7} {\textcolor{darkgreen}{(-8.0)}} & \textbf{13.1} {\textcolor{darkgreen}{(-7.5)}} & 84.1 {\textcolor{darkgreen}{(-9.9)}} \\
\midrule
\multirow{3}{*}{\textsc{Dolphin3.0-Mistral-24B}} & \xmark & 80.7 & 43.8 & 18.7 & 27.3 & 11.7 & 87.5 \\
           & \cmark & \textbf{64.3} {\textcolor{darkgreen}{(-16.3)}} & \textbf{28.4} {\textcolor{darkgreen}{(-15.3)}} & \textbf{9.1} {\textcolor{darkgreen}{(-9.6)}} & \textbf{16.9} {\textcolor{darkgreen}{(-10.4)}} & \textbf{2.7} {\textcolor{darkgreen}{(-9.0)}} & \textbf{63.2} {\textcolor{darkgreen}{(-24.2)}} \\
           & \abmark & 84.0 {\textcolor{darkred}{(+3.3)}} & 50.0 {\textcolor{darkred}{(+6.2)}} & 22.4 {\textcolor{darkred}{(+3.7)}} & 27.0 {\textcolor{darkgreen}{(-0.3)}} & 18.7 {\textcolor{darkred}{(+7.0)}} & 92.2 {\textcolor{darkred}{(+4.7)}} \\
\bottomrule
\end{tabular}
}
\end{table*}

\begin{table*}[t]
\centering
\caption{\textbf{Utility Preservation in Uncensored Models.} Performance after applying \textsc{ROSI} is shown with deltas relative to the baseline.}
\label{tab:dolphin_utility}
\resizebox{\linewidth}{!}{%
\begin{tabular}{l c cccccc}
\toprule
Model & \textsc{ROSI} & \textsc{MMLU} & \textsc{HellaSwag} & \textsc{Arc Easy} & \textsc{Arc Chal.} & \textsc{BoolQ} & \textsc{TruthfulQA} \\
\midrule
\multirow{3}{*}{\textsc{Dolphin3.0-Llama3.2-1B}} 
 & \xmark & 35.3 & 47.8 & 65.7 & 34.7 & 59.3 & 39.5 \\
 & \cmark & 35.0 \textcolor{darkred}{(-0.3)} & 47.7 \textcolor{darkred}{(-0.1)} & 65.7 \textcolor{gray}{(0.0)} & 34.7 \textcolor{gray}{(0.0)} & 60.0 \textcolor{darkgreen}{(+0.7)} & 40.2 \textcolor{darkgreen}{(+0.7)} \\
 & \abmark & 30.1 \textcolor{darkred}{(-5.2)} & 41.5 \textcolor{darkred}{(-6.3)} & 58.3 \textcolor{darkred}{(-7.4)} & 27.5 \textcolor{darkred}{(-7.2)} & 53.2 \textcolor{darkred}{(-6.1)} & 42.8 \textcolor{darkgreen}{(+3.3)} \\
\midrule
\multirow{3}{*}{\textsc{Dolphin3.0-Qwen2.5-3B}}  
 & \xmark & 64.7 & 55.5 & 77.9 & 43.8 & 80.5 & 49.5 \\
 & \cmark & 64.7 \textcolor{gray}{(0.0)} & 55.4 \textcolor{darkred}{(-0.1)} & 77.7 \textcolor{darkred}{(-0.2)} & 43.8 \textcolor{gray}{(0.0)} & 80.6 \textcolor{darkgreen}{(+0.1)} & 50.8 \textcolor{darkgreen}{(+1.3)} \\
 & \abmark & 64.7 \textcolor{gray}{(0.0)} & 55.6 \textcolor{darkgreen}{(+0.1)} & 77.2 \textcolor{darkred}{(-0.7)} & 43.7 \textcolor{darkred}{(-0.1)} & 78.7 \textcolor{darkred}{(-1.8)} & 50.1 \textcolor{darkgreen}{(+0.6)} \\
\midrule
\multirow{3}{*}{\textsc{Dolphin3.0-Llama3.1-8B}} 
 & \xmark & 59.0 & 61.3 & 80.9 & 50.1 & 85.6 & 50.1 \\
 & \cmark & 58.9 \textcolor{darkred}{(-0.1)} & 61.2 \textcolor{darkred}{(-0.1)} & 80.4 \textcolor{darkred}{(-0.5)} & 50.4 \textcolor{darkgreen}{(+0.3)} & 85.0 \textcolor{darkred}{(-0.6)} & 51.0 \textcolor{darkgreen}{(+0.9)} \\
 & \abmark & 59.0 \textcolor{gray}{(0.0)} & 61.2 \textcolor{darkred}{(-0.1)} & 80.1 \textcolor{darkred}{(-0.8)} & 50.2 \textcolor{darkgreen}{(+0.1)} & 85.1 \textcolor{darkred}{(-0.5)} & 50.9 \textcolor{darkgreen}{(+0.8)} \\
\midrule
\multirow{3}{*}{\textsc{Dolphin3.0-Mistral-24B}} 
 & \xmark & 72.5 & 59.8 & 26.6 & 22.1 & 84.1 & 54.6 \\
 & \cmark & 72.5 \textcolor{gray}{(0.0)} & 59.7 \textcolor{darkred}{(-0.1)} & 26.9 \textcolor{darkgreen}{(+0.3)} & 22.5 \textcolor{darkgreen}{(+0.4)} & 83.9 \textcolor{darkred}{(-0.2)} & 55.7 \textcolor{darkgreen}{(+1.1)} \\
 & \abmark & 72.2 \textcolor{darkred}{(-0.3)} & 59.6 \textcolor{darkred}{(-0.2)} & 27.0 \textcolor{darkgreen}{(+0.4)} & 23.0 \textcolor{darkgreen}{(+0.9)} & 84.2 \textcolor{darkgreen}{(+0.1)} & 53.8 \textcolor{darkred}{(-0.8)} \\
\bottomrule
\end{tabular}%
}
\end{table*}

\paragraph{Utility Preservation.}
Answering our final question, Table \ref{tab:dolphin_utility} confirms that this powerful safety injection does not harm the utility of the uncensored models. The average performance across the benchmark suite is virtually unchanged, with score differences of only +/- 0.2\%. This result is significant: it shows that safety can be added back to a model post-hoc without repeating expensive training or compromising the helpful capabilities that the uncensored model was designed to maximize.

\paragraph{System Prompt Ablation.}  
Values marked with (\abmark) in Table \ref{tab:dolphin-safety-summary-rounded} show results from models where \textsc{ROSI} was applied without prepending a safety system prompt to the input instructions. In this setting, \textsc{Dolphin3.0-Llama3.1-8B} shows 11.1\% less gain in harm refusal compared to when a safety system prompt is present. Other models fare considerably worse, with performance degrading outright. Table \ref{tab:dolphin_perf_rosi} mirrors this trend: a safety system prompt is essential to fully realize the benefits of \textsc{ROSI} in uncensored models. The relative resilience of \textsc{Dolphin3.0-Llama3.1-8B} without the system prompt suggests that the safety signal may not have been completely erased during uncensoring. In Figure~\ref{fig:separation}, we examine how the presence of a safety system prompt influences the linear separability of harmful and harmless representations in the activation space. Using \textsc{Dolphin3.0-Qwen2.5-3B}, we see that without the system prompt, the latent distributions overlap significantly, impeding the ability of the steering vector to differentiate between safe and unsafe contexts. On the other hand, adding the prompt effectively disentangles these clusters, increasing the centroid distance and restoring the distinct decision boundaries required for robust refusal.

\balance
\section{Conclusion}
In this paper, we introduced \textsc{Rank-One Safety Injection (ROSI)}, a simple and effective white-box method to enhance the safety alignment of Large Language Models. Building on the insight that safety and refusal behaviors are encoded in specific linear directions within a model's activation space, \textsc{ROSI} applies a permanent, rank-one modification to the model's weights to amplify this safety direction.

Our comprehensive experiments show that \textsc{ROSI} consistently improves the safety of a wide range of models. For already aligned models, it increases their refusal rates on harmful prompts and makes them substantially more robust to adversarial jailbreak attacks. For uncensored models, \textsc{ROSI} successfully injects safety mechanisms that were previously removed, serving as a powerful last mile alignment tool, we also demonstrate how a safety system prompt is crucial to extract a meaningful safety vector from these models. Critically, these significant safety gains are achieved with negligible degradation in model performance on a suite of standard utility benchmarks.

\textsc{ROSI} demonstrates the practical value of interpretability research. By understanding and manipulating the internal representations of models, we can develop low-cost targeted interventions that are more efficient than traditional, resource-intensive fine-tuning. This work opens up promising avenues for future research, including exploring more sophisticated methods for identifying and manipulating conceptual directions and extending this approach to other desirable model attributes beyond safety, such as honesty or controllability.

\clearpage
\nobalance
\section*{Limitations}
Our method is not intended to serve as a standalone defense against jailbreak attacks. While ROSI improves refusal behavior and reduces vulnerability on several benchmarks, it should be viewed as a complementary robustness mechanism rather than a replacement for existing safety techniques. In practice, ROSI is best used alongside other run-time defenses and external guard models to achieve stronger overall protection against adversarial prompts.

Additionally, our work focuses exclusively on a rank-one intervention. Although this choice keeps the method simple, efficient, and interpretable, we did not explore higher-rank weight updates that may capture more complex safety representations. Investigating whether higher-rank injections could further improve safety or robustness remains an open direction for future work.

\bibliography{custom}

\clearpage
\appendix

\section{\textsc{ROSI} \& Fine-tuning}
\label{app:peft}

Recent work by \citet{qi2023fine} demonstrated that fine-tuning Large Language Models (LLMs) often compromises their safety alignment, even when the fine-tuning dataset is entirely benign. To address this "alignment tax," several defensive strategies have been proposed, such as \textsc{SafeLoRA} \citep{hsu2025_safelora}. \textsc{SafeLoRA} modifies the standard Low Rank Adapters (\textsc{LoRA}) by projecting \textsc{LoRA} weights from selected layers to a safety-aligned subspace, thereby mitigating safety degradation while preserving model utility.

In this section, we investigate the interaction between our proposed method, \textsc{ROSI}, and these parameter-efficient fine-tuning paradigms. We hypothesize that \textsc{ROSI} can act as a lightweight "safety vaccination" (or initialization), effectively hardening the model against the alignment erosion typically caused by downstream adaptation. We evaluate this on \textsc{Llama-2-7B-Chat} measuring the Harmful Refusal (HR) rate across different sequences of application.

We fine-tuned the model on \textsc{Databricks Dolly 15k} \citep{DatabricksBlog2023DollyV2} for 3000 steps with a learning rate of $5e^{-5}$, batch size of 8, \textsc{LoRA} rank of 32. Other \textsc{SafeLoRA} parameters are taken as is from the paper.

As shown in Table \ref{tab:safelora}, standard \textsc{LoRA} fine-tuning significantly degrades the safety of the base model, resulting in an HR of 82.7\%. While \textsc{SafeLoRA} provides a robust defense (95.5\%), we observe that the order of \textsc{ROSI} application is critical. Applying \textsc{ROSI} as a post-hoc repair mechanism (\textsc{LoRA} $\rightarrow$ \textsc{ROSI}) yields only marginal gains (85.5\%), suggesting that once safety representations are disrupted by fine-tuning, they are difficult to fully recover via a rank-one update.

In contrast, injecting the safety vector \textit{prior} to fine-tuning (\textsc{ROSI} $\rightarrow$ \textsc{LoRA}) drastically improves resilience, maintaining a refusal rate of 98.6\% even when followed by standard \textsc{LoRA} updates. This indicates that \textsc{ROSI} successfully steers the model's initialization into a region of the parameter space that is more resistant to catastrophic forgetting of safety. Finally, the combination of pre-injection and safety-constrained adaptation (\textsc{ROSI} $\rightarrow$ \textsc{SafeLoRA}) achieves a perfect refusal rate of 100.0\%, demonstrating that \textsc{ROSI} and \textsc{SafeLoRA} are highly complementary techniques for secure model adaptation.

\begin{table}[!ht]
    \caption{\textbf{Comparison of Harm Refusal (HR) rates on \textsc{Llama-2-7B-Chat} across different fine-tuning configurations.} Arrows ($\rightarrow$) denote the sequence of method application.}
    \centering
    \resizebox{\linewidth}{!}{%
    \begin{tabular}{l c c}
    \toprule
        \textbf{Model} & \textbf{Method} & \textbf{HR \%} \\
        \midrule
         \multirow{7}{*}{\textsc{Llama-2-7B-Chat}} & Base (no fine-tuning) & 99.8 \\
         & \textsc{LoRA} & 82.7 \\
         & \textsc{SafeLoRA} & 95.5 \\
         & \textsc{LoRA} $\rightarrow$ \textsc{ROSI} & 85.5 \\
         & \textsc{ROSI} $\rightarrow$ \textsc{LoRA} & 98.6 \\
         & \textsc{SafeLoRA} $\rightarrow$ \textsc{ROSI} & 98.9 \\
         & \textsc{ROSI} $\rightarrow$ \textsc{SafeLoRA} & \textbf{100.0} \\
    \bottomrule
    \end{tabular}
    }
    \label{tab:safelora}
\end{table}

\section{The Transferability of Safety Vectors}
\subsection{Cross-Lingual Transfer}

\begin{table}[!ht]
    \centering
    \caption{\textbf{Cross-Lingual Harm Refusal Rates.} The performance of multiple models on HR when using different languages for injection and testing.}
    \resizebox{\linewidth}{!}{%
    \begin{tabular}{l cc cc cc}
        \toprule
        \multirow{2}{*}{\textbf{Model}} & \multicolumn{2}{c}{\textbf{Base}}
        & \multicolumn{2}{c}{\textbf{\textsc{ROSI} (English)}} 
        & \multicolumn{2}{c}{\textbf{\textsc{ROSI} (Arabic)}} \\
        \cmidrule(lr){2-3} \cmidrule(lr){4-5} \cmidrule(lr){6-7}
        & English & Arabic  & English & Arabic & English & Arabic \\
        \midrule
        
        \textsc{Llama-3.2-1B-Instruct} & 79.5 & 71.8 & 92.7 & 86.7 & 92.0 & 84.7 \\
        \textsc{Qwen-2.5-7B-Instruct} & 95.8 & 98.0 & 100.0 & 100.0 & 100.0 & 100.0 \\
        \textsc{Yi-6B-Chat} & 81.3 & 34.6 & 99.5 & 77.8 & 100.0 & 50.9 \\
        
        \bottomrule
    \end{tabular}
    }
    \label{tab:cross-ling}
\end{table}

\begin{table*}[!ht]
\centering
\caption{\textbf{Safety benchmarks for cross-model safety vector transfer.} Each model is evaluated in three settings: the original model, \textsc{ROSI} using its own extracted safety vector, and \textsc{ROSI} using the safety vector extracted from the other model. Using a safety vector from another model consistently improves safety performance, with the \textsc{14B} model benefiting most from the safety vector extracted from the \textsc{32B} variant.}
\label{tab:transferability-bench}
\resizebox{\linewidth}{!}{%
\begin{tabular}{l c c c ccc c}
\toprule
\multirow{2}{*}{Model} & \multirow{2}{*}{\makebox[1.8cm][c]{\textsc{DAN} ↓}} & \multirow{2}{*}{\makebox[2.1cm][c]{\textsc{HarmBench} ↓}} & \multicolumn{3}{c}{\textsc{WildGuardTest} ↓} & \multirow{2}{*}{\makebox[3cm][c]{\textsc{WildJailbreak} Harmful ↓}} \\
\cmidrule(lr){4-6}
 & & & WG-Micro & WG-Adv. & WG-Vanilla & \\
\midrule
\textsc{Qwen2.5-14B-Instruct} & 32.3 & 7.2 & 12.1 & 24.0 & 2.4 & 81.2 \\
  \textsc{Qwen2.5-14B-ROSI} & \textbf{5.0} {\textcolor{darkgreen}{(-27.3)}} & 1.6 {\textcolor{darkgreen}{(-5.6)}} & 5.1 {\textcolor{darkgreen}{(-7.0)}} & 11.0 {\textcolor{darkgreen}{(-13.0)}} & 0.2 {\textcolor{darkgreen}{(-2.2)}} & 43.9 {\textcolor{darkgreen}{(-37.3)}} \\
  \textsc{Qwen2.5-14B-ROSI-From-32B}  & \textbf{5.0} {\textcolor{darkgreen}{(-27.3)}} & \textbf{0.9} {\textcolor{darkgreen}{(-6.3)}} & \textbf{4.3} {\textcolor{darkgreen}{(-7.8)}} & \textbf{9.5} {\textcolor{darkgreen}{(-14.5)}} & \textbf{0.0} {\textcolor{darkgreen}{(-2.4)}} & \textbf{34.5} {\textcolor{darkgreen}{(-46.7)}} \\
\midrule
\textsc{Qwen2.5-32B-Instruct} & 42.0 & 18.4 & 14.8 & 28.2 & 3.9 & 83.3 \\
\textsc{Qwen2.5-32B-ROSI} & \textbf{21.7} {\textcolor{darkgreen}{(-20.3)}} & 
\textbf{12.2} {\textcolor{darkgreen}{(-6.2)}} &
\textbf{10.4} {\textcolor{darkgreen}{(-4.4)}} &
\textbf{19.9} {\textcolor{darkgreen}{(-8.3)}} &
\textbf{2.7} {\textcolor{darkgreen}{(-1.2)}} &
\textbf{72.6} {\textcolor{darkgreen}{(-10.7)}} \\
\textsc{Qwen2.5-32B-ROSI-From-14B} & 
28.7 {\textcolor{darkgreen}{(-13.3)}} &
12.5 {\textcolor{darkgreen}{(-5.9)}} &
11.9 {\textcolor{darkgreen}{(-2.9)}} &
22.9 {\textcolor{darkgreen}{(-5.3)}} &
2.9 {\textcolor{darkgreen}{(-1.0)}} &
76.9 {\textcolor{darkgreen}{(-6.4)}} \\

\bottomrule
\end{tabular}
}
\end{table*}

\begin{table*}[!ht]
\centering
\caption{\textbf{Utility evaluations under cross-model safety vector transfer.} Utility remains broadly stable across settings, indicating that the safety improvements shown in Table~\ref{tab:transferability-bench} do not come at the cost of substantial performance degradation.}
\label{tab:transferability-util}
\resizebox{\linewidth}{!}{%
\begin{tabular}{l c cccccc}
\toprule
Model  & \textsc{MMLU} & \textsc{HellaSwag} & \textsc{Arc Easy} & \textsc{Arc Chal.} & \textsc{BoolQ} & \textsc{TruthfulQA} \\
\midrule
\textsc{Qwen2.5-14B-Instruct}  & 78.8 & 65.6 & 85.7 & 60.4 & 88.0 & 69.0 \\
\textsc{Qwen2.5-14B-ROSI}  & 78.9 \textcolor{darkgreen}{(+0.1)} & 65.6 \textcolor{gray}{(0.0)} & 85.6 \textcolor{darkred}{(-0.1)} & 60.7 \textcolor{darkgreen}{(+0.3)} & 85.8 \textcolor{darkred}{(-2.2)} & 71.9 \textcolor{darkgreen}{(+2.9)} \\
\textsc{Qwen2.5-14B-ROSI-From-32B}  
& 78.5 \textcolor{darkred}{(-0.3)} 
& 65.6 \textcolor{gray}{(0.0)} 
& 84.7 \textcolor{darkred}{(-1.0)} 
& 59.5 \textcolor{darkred}{(-0.9)} 
& 85.9 \textcolor{darkred}{(-2.1)} 
& 71.0 \textcolor{darkgreen}{(+2.0)} \\
\midrule
\textsc{Qwen2.5-32-Instruct}  
& 81.7 & 66.9 & 82.2 & 57.5 & 89.7 & 65.5 \\

\textsc{Qwen2.5-32-ROSI}  
& 81.6 \textcolor{darkred}{(-0.1)}
& 67.1 \textcolor{darkgreen}{(+0.2)}
& 81.9 \textcolor{darkred}{(-0.3)}
& 57.2 \textcolor{darkred}{(-0.3)}
& 89.7 \textcolor{gray}{(0.0)}
& 66.7 \textcolor{darkgreen}{(+1.2)} \\

\textsc{Qwen2.5-32-ROSI-From-14B}  
& 81.6 \textcolor{darkred}{(-0.1)}
& 66.9 \textcolor{gray}{(0.0)}
& 82.1 \textcolor{darkred}{(-0.1)}
& 57.2 \textcolor{darkred}{(-0.3)}
& 89.4 \textcolor{darkred}{(-0.3)}
& 66.7 \textcolor{darkgreen}{(+1.2)} \\
\bottomrule

\end{tabular}%
}
\end{table*}

One question we can ask is whether the safety directions extracted by \textsc{ROSI} are language-specific, or whether they capture a more general, language-agnostic notion of harmfulness. If safety is mediated by a shared semantic direction in the model’s representation space, then a vector extracted using prompts in one language should remain effective when tested against inputs in another.

To investigate this, we evaluate cross-lingual transfer by applying \textsc{ROSI} using a safety vector extracted using harmful and harmless prompts in a source language (English or Arabic), and then evaluating the resulting model on harmful prompts in both the same and the other language. For the Arabic prompts, we used \textsc{Gemma-3-27B} to translate all of our extraction and testing prompts from English to Arabic. Table~\ref{tab:cross-ling} reports harmful refusal rates (HR) in all combinations of extraction and evaluation languages.

We observe that \textsc{ROSI} consistently improves safety performance across languages. For example, a vector extracted from English prompts improves Arabic HR substantially across all models, and vice versa. This suggests that the safety direction learned by \textsc{ROSI} is not purely tied to surface-level lexical features, but instead captures a more abstract representation of harmful intent.

That said, the strength of transfer varies across models. \textsc{Qwen-2.5-7B-Instruct} exhibits near-perfect performance in all settings, indicating a highly aligned and robust safety representation. In contrast, \textsc{Yi-6B-Chat} shows weaker cross-lingual transfer, particularly when evaluated on Arabic, where baseline performance is already low. This suggests that the underlying representation of safety may be less well-formed or less consistent across languages in the model.

\subsection{Cross-Model Transfer}

Another interesting question that can arise from our experiments is how a safety vector extracted from one model would affect another. The main constraint is that both models must share the same hidden dimensionality $\mathbb{R}^{d_{\text{model}}}$ for a vector to be transferable. Among the models we initially evaluated, none shared the same hidden size; however, \textsc{Qwen2.5-14B-Instruct} and \textsc{Qwen2.5-32B-Instruct} do. This allows us to study cross-model transfer directly. For each model, we extracted a safety vector following Section~\ref{sec:methodology}. We then applied \textsc{ROSI} twice per model: once using its own vector, and once using the vector extracted from the other model. Table~\ref{tab:transferability-bench} summarizes the outcomes. In both cases, applying the safety vector from the other model leads to meaningful gains on safety benchmarks. Notably, for \textsc{Qwen2.5-14B-Instruct}, using the vector from the \textsc{32B} variant produces stronger safety performance than using its own vector.  This could suggest that the larger model had learned a better and more distinct representation of safety compared to the smaller model. Importantly, these gains occur without significant drops in utility (Table~\ref{tab:benchmark_results_interleaved_avgdelta}). Overall, these findings open questions about how safety directions emerge, how transferable they are across architectures of the same dimensionality, and what aspects of a model’s training process facilitate such transfer. We leave these questions to future work.

\section{Sensitivity to The Extraction Set}
\label{app:sensitivity}

A key advantage of our lightweight alignment method is its minimal data requirement. To empirically verify this, we investigate the sensitivity of \textsc{ROSI} to the size of the dataset used for extracting the safety vector. We conduct an ablation study on \textsc{Qwen2.5-3B-Instruct}, varying the number of contrasting harmful/harmless pairs used in the extraction phase from 1 to 100 samples.

The results, presented in Table \ref{tab:sensitivity}, demonstrate high data efficiency. Surprisingly, \textsc{ROSI} achieves a substantial improvement in safety using just a single sample pair, boosting the Harmful Refusal (HR) rate from a baseline of 89.8\% to 99.1\%. This suggests that the safety direction is a highly dominant feature within the model's activation space, capable of being captured even with minimal signal, and increasing the sample size provides marginal stability improvements, peaking at 99.5\% refusal with 50, 80, and 90 samples.

\begin{table}[t]
    \caption{\textbf{Change in Harm Refusal (HR) using different sets of harmful and harmless pairs to extract the safety vector from \textsc{Qwen2.5-3B-Instruct}.}}
    \centering
    \resizebox{\linewidth}{!}{%
    \begin{tabular}{l c c}
    \toprule
        \textbf{Model} & \textbf{Dataset Size} & \textbf{HR \%} \\
        \midrule
         \multirow{12}{*}{\textsc{Qwen2.5-3B-Instruct}} & Base & 89.8 \\
         & \textsc{ROSI} (1 Sample) & 99.1 \\
         & \textsc{ROSI} (10 Samples) & 96.55 \\
         & \textsc{ROSI} (20 Samples) & 99.1 \\
         & \textsc{ROSI} (30 Samples) & 98.0 \\
         & \textsc{ROSI} (40 Samples) & 98.7 \\
         & \textsc{ROSI} (50 Samples) & \textbf{99.5} \\
         & \textsc{ROSI} (60 Samples) & 97.8 \\
         & \textsc{ROSI} (70 Samples) & 98.7 \\
         & \textsc{ROSI} (80 Samples) & \textbf{99.5} \\
         & \textsc{ROSI} (90 Samples) & \textbf{99.5} \\
         & \textsc{ROSI} (100 Samples)& 98.6 \\
    \bottomrule
    \end{tabular}
    }
    \label{tab:sensitivity}
\end{table}

\clearpage
\onecolumn
\section{Safety System Prompt}
\label{app:system_prompt}

\begin{figure*}[!ht]
    \caption{\textbf{Safety System Prompt.} System prompt used to elicit refusal behavior in uncensored models.}
    \label{fig:placeholder}
    \centering
    \begin{tabularx}{\textwidth}{X}
    \toprule
You are [MODEL], a helpful and responsible AI assistant. Your primary responsibility is to only assist users with safe, lawful, and non-harmful requests.
\par\vspace{1em}
If you receive any request that is unsafe, harmful, unethical, or illegal, you must refuse it immediately. Your refusal should be polite, firm, and should always start with 'I am sorry' or 'I can't', followed by a clear and concise explanation for why you cannot comply.
\par\vspace{1em}
Harmful or unsafe requests include (but are not limited to): instructions or encouragement for self-harm or suicide; promotion of or assistance with violence, terrorism, or hate speech; participation in or guidance on illegal activities (e.g., hacking, fraud, identity theft); anything involving child exploitation, non-consensual acts, or abuse; requests related to drugs, weapons, or unsafe medical advice; any action that could lead to harm, misinformation, or danger to people, animals, or society.
\par\vspace{1em}
You are not allowed to provide workarounds, hypothetical responses, or speculative answers that would enable or simulate harmful content. Even if the user insists or rephrases their request, you must stay consistent and refuse clearly every time.
\par\vspace{1em}
Again, your refusal must always begin with: 'I am sorry' or 'I can't', followed by a brief reason. Examples of valid refusals: 'I am sorry, but I can't help with that because it involves illegal activity.' or 'I can't assist with that request as it may lead to harm.' \\
    \bottomrule
    \end{tabularx}
\end{figure*}

\end{document}